\documentclass[10pt,twocolumn,letterpaper]{article}

\usepackage{cvpr}
\usepackage{times}
\usepackage{epsfig}
\usepackage{graphicx}
\usepackage{amsmath}
\usepackage{amssymb}
\usepackage{cite}
\usepackage{amsmath,amssymb,amsfonts}
\usepackage{algorithmic}
\usepackage{textcomp}
\usepackage{booktabs}
\usepackage{bm}
\usepackage{multirow}
\usepackage{float}
\usepackage{algorithm}
\usepackage{xspace}

\usepackage[pagebackref=true,breaklinks=true,letterpaper=true,colorlinks,bookmarks=false]{hyperref}

\begin{document}


\title{Few-Shot Learning: Expanding ID Cards Presentation Attack Detection to Unknown ID Countries}

\author{Alvaro S. Rocamora and Juan M. Espín\\
Facephi company, Spain\\
{\tt\small asanchezrocamora, jmespin@facephi.com}
\and
Juan E. Tapia\\
Hochschule Darmstadt\\
da/sec-Biometrics and Internet Security Research Group, Germany\\
{\tt\small juan.tapia-farias@h-da.de}
}

\maketitle
\thispagestyle{empty}

\begin{abstract}
This paper proposes a Few-shot Learning (FSL) approach for detecting Presentation Attacks on ID Cards deployed in a remote verification system and its extension to new countries. 
Our research analyses the performance of Prototypical Networks across documents from Spain and Chile as a baseline and measures the extension of generalisation capabilities of new ID Card countries such as Argentina and Costa Rica. Specifically targeting the challenge of screen display presentation attacks. By leveraging convolutional architectures and meta-learning principles embodied in Prototypical Networks, we have crafted a model that demonstrates high efficacy with Few-shot examples. This research reveals that competitive performance can be achieved with as Few-shots as five unique identities and with under 100 images per new country added. This opens a new insight for novel generalised Presentation Attack Detection on ID cards to unknown attacks.

\end{abstract}

\section{Introduction}
\label{sec:introduction}

In the dynamic environment of digital security, the challenge of identifying fraudulent activities, especially in the remote verification system, is more critical than ever. This is even truer if we detect attacks on ID cards from different countries. Among these emerging threats, screen display attacks on identity documents, such as ID Cards, stand out as a particularly insidious challenge because of the rising quality of cameras in smartphones. These attacks, where fraudsters present digital replicas of genuine documents on screens to deceive verification systems, demand innovative detection and prevention strategies.

In scenarios where acquiring genuine ID Cards poses significant challenges due to privacy and safety concerns, the potential of meta-learning approaches becomes particularly convincing \cite{TIAN2022203, survey_FSL}. The driving force behind Zero and Few Shot learning promises to greatly reduce the need for extensive data collection. Instead of relying on thousands of labelled ID Card examples to achieve satisfactory performance in a new ID Card country, Zero-shot or Few-shot aims to streamline the model training process. This not only reduces the data loading gathering but also cuts down on computational costs and the time to train models.

Further, the scarcity of databases with examples of screen display ID Card attacks significantly hampers the development and testing of Few Shot Learning (FSL) models that could effectively address this specific challenge. Most publicly available datasets for research are limited and often consist of synthetic or low-quality artificially generated ID Cards rather than real-world examples of screen display fraud \cite{midv-2019, midv-holo}. This limitation not only curtails innovation but also diminishes the effectiveness of meta-learning models by restricting their exposure to a broad spectrum of attack methodologies.

Moreover, the reliance on synthetic datasets introduces a significant hurdle. Such datasets lack the complexity and subtleties of real-world forgeries, including the unique artefacts introduced by screen displays, such as moiré patterns, brightness and contrast issues, and other digital anomalies \cite{Benalcazar, markham2023openset}. This gap between synthetic training data and the realities of screen display attacks could result in a notable performance disparity when these models are deployed in real remote verification scenarios. However, getting high-quality ID Card images (Face and Text together) is still an open challenge.

Motivated by the previous challenges, this article develops a solution to extend Presentation Attack Detection (PAD) to a new country by using a reduced number of images and applying FSL to address screen display presentation attacks on ID Cards specifically. 

FSL holds the key to unlocking insights in domains where data is scarce or difficult to obtain, navigating through the constraints of privacy and safety with ease. By empowering models to generalise from a handful of examples, FSL offers a viable pathway to leveraging sensitive or hard-to-access data effectively, marking a significant stride towards more efficient and adaptable machine learning methodologies.

Our focus is on solving the challenge of limited ID Card data availability. Through this exploration, we aim to highlight FSL's potential in transforming the fight against digital identity fraud and to pave the way for future research and development to enhance the security of identity verification systems against the sophisticated threat of screen display attacks.

This paper develops the use of FSL techniques to identify and counteract screen display presentation attacks on ID Cards from 4 different countries. The FSL generalising to new cases could be revolutionary for combating such sophisticated remote fraud techniques and its extension to new ID Cards. However, a survey of current literature indicates a notable lack of FSL applications specifically designed to tackle Presentation Attack Detection and screen display attacks on ID Card in the context of remote verification systems.

The main contributions of this paper are:
\begin{itemize}
  \item Enhances detection capabilities for display attacks on screens using minimal samples (Few-shot), demonstrating effective learning from a few examples.
    \item Reduces the collection and creation of large datasets of sensitive information, like ID Cards, by employing FSL based on a Prototypical Network, thus minimising data requirements.
  \item Our findings show adaptability and robust performance across diverse geographical contexts, including Spain, Chile, Argentina, and Costa Rica.  
  \item Implements a training methodology based on EfficientNetV2-B0 that mirrors real-world scenarios, enabling a fast adaptation to new tasks.

\end{itemize}

The rest of the article is organised as follows: Section~\ref{sec:related} summarises the related works on FSL. Section~\ref{sec:databases} describes and depicts examples of images and countries ID Cards. The metrics used for the evaluation and the proposed methods based on the prototypical network are in Sections ~\ref{sec:metrics} and ~\ref{sec:method}, respectively. The experimental framework and results of this work are then presented in Section~\ref{sec:exp-result}. We conclude the article in Section~\ref{sec:conclu}.

\section{Related work}

\label{sec:related}
FSL represents a pivotal shift in the paradigm of machine learning, addressing the challenge of learning from minimal data. Inspired by the remarkable ability of human cognition to make robust decisions and learn from very few examples, this technique mimics this adaptability in machine learning models \cite{fsl-paper-1, survey_FSL}.

This technique stands as a transformative approach within the realm of machine learning, depicting the application of meta-learning principles to the challenges of supervised learning \cite{survey_FSL}. The essence of FSL is deeply rooted in meta-learning or ``learning to learn", a concept that extends the capabilities of traditional supervised learning models by equipping them with the ability to generalise from limited data based on prior knowledge and learning experiences \cite{TIAN2022203}.

Meta-learning emerges as a powerful paradigm in machine learning, focusing on designing models that can learn new tasks efficiently with minimal supervision \cite{tapia2022-fewshot, survey_FSL}. At its core, meta-learning aims to create systems that improve their learning algorithms over time, accumulating knowledge that can be effectively applied to novel problems. This is achieved by exposing the model to a wide variety of learning tasks during the training phase, thereby enabling it to identify patterns, strategies, and principles that are transferable across tasks. The model essentially learns the optimal way to adapt its parameters to new, previously unseen tasks based on its accumulated learning experience.

Two standout strategies with meta-learning concepts in this domain are highlighted in the state-of-the-art such as Prototypical Networks \cite{prototypical-network} and Relational Networks \cite{relational-network}, each addressing the challenges of FSL through distinct meta-learning lenses. 

Prototypical Networks fall under the metric-based meta-learning category, where learning a ``similarity metric" is the strategy. This approach involves classifying new examples by comparing them to prototype representations of each class, calculated as the mean vector of their features in a learned space \cite{prototypical-network}. The idea is straightforward yet powerful: objects of the same class should cluster around their prototype in the feature space, making classification a matter of proximity.

On the other hand, Relational Networks are designed to adapt to new tasks through static models and by analysing and understanding the relationships and interactions between data points \cite{relational-network}. By focusing on the dynamic aspects of data relationships, Relational Networks are generalised from minimal data, making them very suitable for presentation attack detection on ID Cards, enabling them to tackle new challenges with an informed perspective based on learned relations.

\subsection{Prototypical Networks on FSL}
Prototypical Networks focus on classifying a small number of examples (ID Cards) by generating a prototype, a representative example, for each class \cite{prototypical-network}. For instance, if you have a ``Support set" composed of $N$-labelled examples. Each of these examples has embeddings and a corresponding label indicating its class out of a total of $K$ possible classes.

\begin{figure}[ht]
\centering
    \includegraphics[width=0.8\columnwidth]{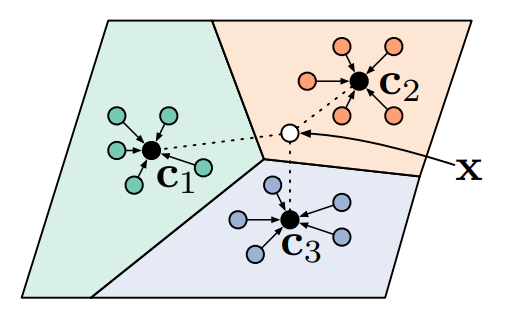}
    \caption{Example of Few-shot prototypes $c_{k}$ are computed as the mean of embedded support examples for each class \cite{prototypical-network}.}
    \label{fig:prototypical-net}
\end{figure}

The process starts by transforming these embeddings into a new space using a function that is designed to highlight the features most relevant to classifying the examples. In this transformed space, the prototype of each class is calculated simply as the average of all the embeddings belonging to that class. This means if you are looking at a particular class (bona fide or screen display attack, in our case), its prototype is the central point that best represents all the examples of that class in the new space $c_{k}$, as is shown in Figure~\ref{fig:prototypical-net}.

When it comes to classifying a new example (an ID Card from a new country), referred to as a query point and represented as $X$ in Figure~\ref{fig:prototypical-net}, the Prototypical Network evaluates how close or far this new example is from each class's prototype using a distance measure, such as Euclidean or Cosine. Based on these distances, the model predicts the class of the new example by considering which class prototype it is closest to.

The model learns to do this efficiently through training, where it goes through numerous epochs, each involving a random selection of classes and examples that serve as both the ``Support set" and ``Query points". Continuously adjusting the parameters defines how the embeddings are transformed. The goal is to maximise the model's ability to correctly predict the classes of new examples based on their proximity to the class prototypes. This training process is aimed at refining the model's ability to generate accurate prototypes and make reliable class predictions, all while using a minimal amount of data.

\section{Databases}
\label{sec:databases}
For the application of FSL to detect presentation attacks on ID Cards depends on diverse and representative datasets. Current open-access datasets like MIDV 500 \cite{MIDV-500}, MIDV 2019 \cite{midv-2019}, MIDV 2020 \cite{MIDV2020AC} and DLC 2021 \cite{dlc2021}, despite offering a rich amount of country representations and document types, fall short due to their limited number of unique user identities and few examples of screen displays attacks on ID Cards. Conversely, private datasets are not available to compare the results. This fundamental limitation undermines the potential of FSL models to accurately learn and generalise across the wide variability inherent in ID Cards. The depth of unique user data as a starting point is crucial for teaching models to discern between bona fide and fake documents, a complicated task because of the narrow subject base of these datasets. Table~\ref{tab:survey-table} shows a summary of the most relevant database in this field.

\begin{table}[H]
\scriptsize
\centering
\caption{Relevant databases summary selection.}
\label{tab:survey-table}
\begin{tabular}{lcccc}
\toprule
\textbf{Ref.} & \textbf{Dataset} & \textbf{Images} & \textbf{Users} & \textbf{Models} \\ \midrule
\cite{RajaR022}   &  Private  & 104,882   &   886         &     \begin{tabular}[c]{@{}c@{}}Modified \\ DenseNet121 \end{tabular} \\  \midrule
\cite{gonzalez2021hybrid}   &  Private  &     54,980     &    5,000   &  \begin{tabular}[c]{@{}c@{}} MobileNet \\ BasicNet \end{tabular}           \\  \midrule
\cite{gonzalez2023improving}   &  Private  &     190,000     &    16,000   &  \begin{tabular}[c]{@{}c@{}} MobileNet \\ BasicNet \end{tabular}           \\  \midrule
\cite{Benalcazar}           &  Private   & 38,477 &   9,286    &  \begin{tabular}[c]{@{}c@{}} GANs \\ MobileNetV2 \end{tabular}       \\  \midrule
\cite{markham2023openset}   & \begin{tabular}[c]{@{}c@{}}MIDV-2020 \\ DLC-2021\end{tabular}    &  70,050   &   1,050         &  \begin{tabular}[c]{@{}c@{}} GANs \\ MobileNetV2 \end{tabular}  

\\ \bottomrule
\end{tabular}
\end{table}

Given the significant limitations of public datasets for studying FSL applications, we have opted to create a private dataset only for research purposes that accomplish the previously explained initial condition. 

This private dataset was created with digital user document templates generated in-house to simulate a wide array of ID Cards from countries such as Spain (ESP), Chile (CHL), Argentina (ARG), and Costa Rica (CRI). 

\begin{figure}[]
\centering
\subfloat[ESP ID Card Template]{
  \includegraphics[width=35mm]{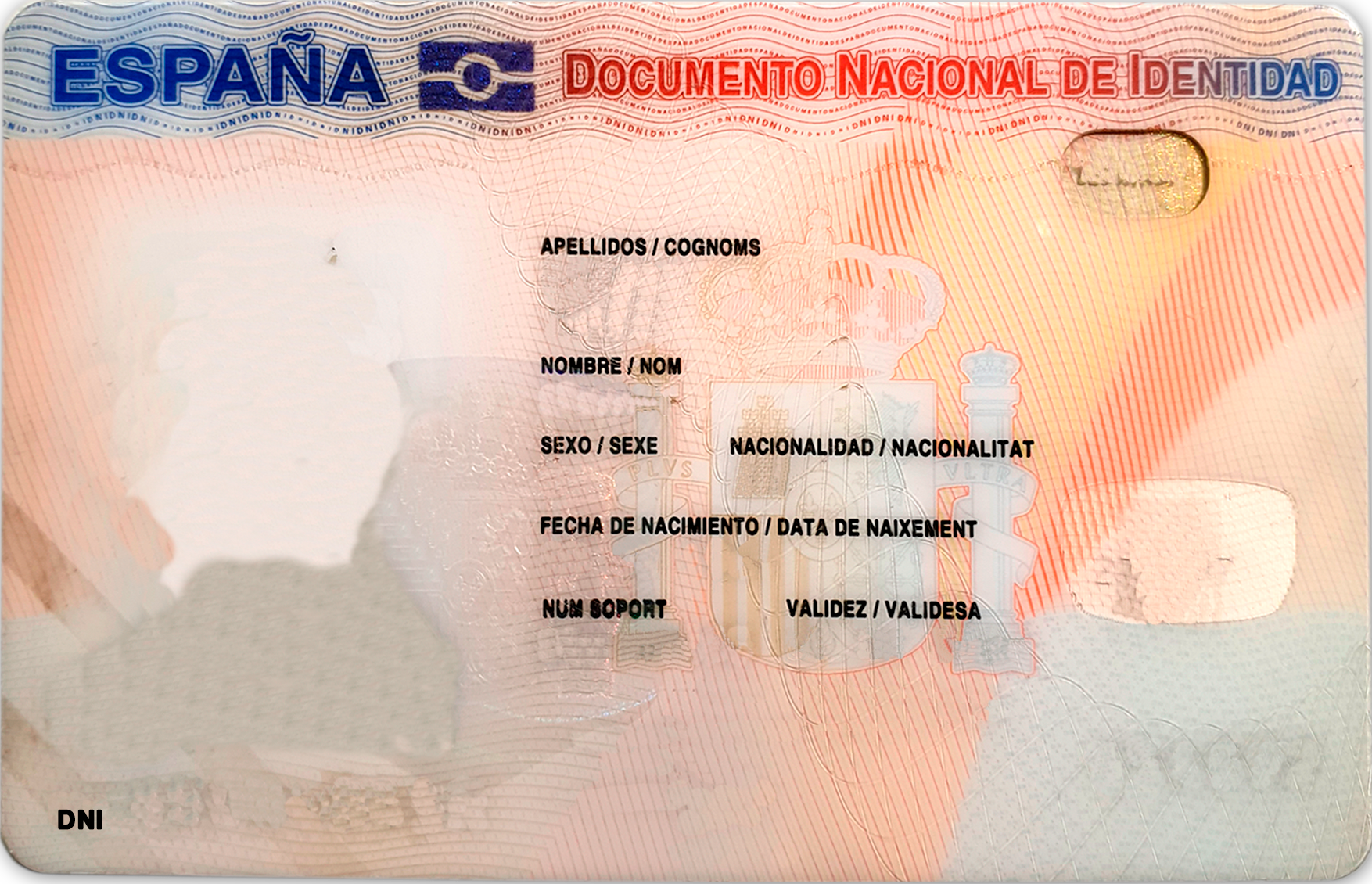}
  \label{fig:empty-esp}
}
\subfloat[CHL ID Card Template]{
  \includegraphics[width=35mm]{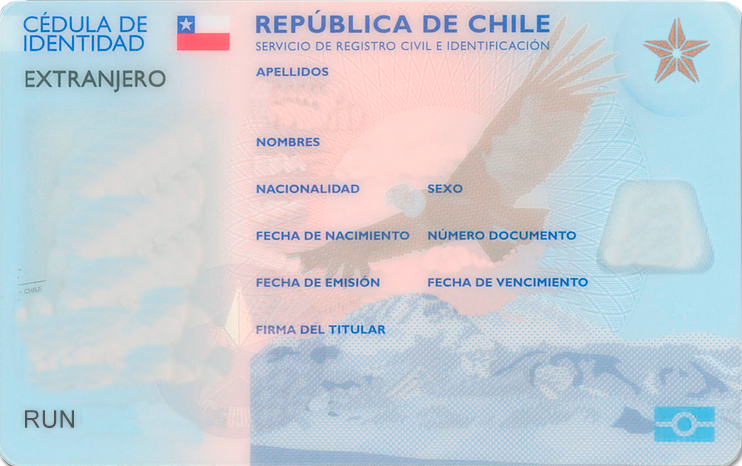}\label{fig:empty-chl}
}

\subfloat[ARG ID Card Template]{
  \includegraphics[width=35mm]{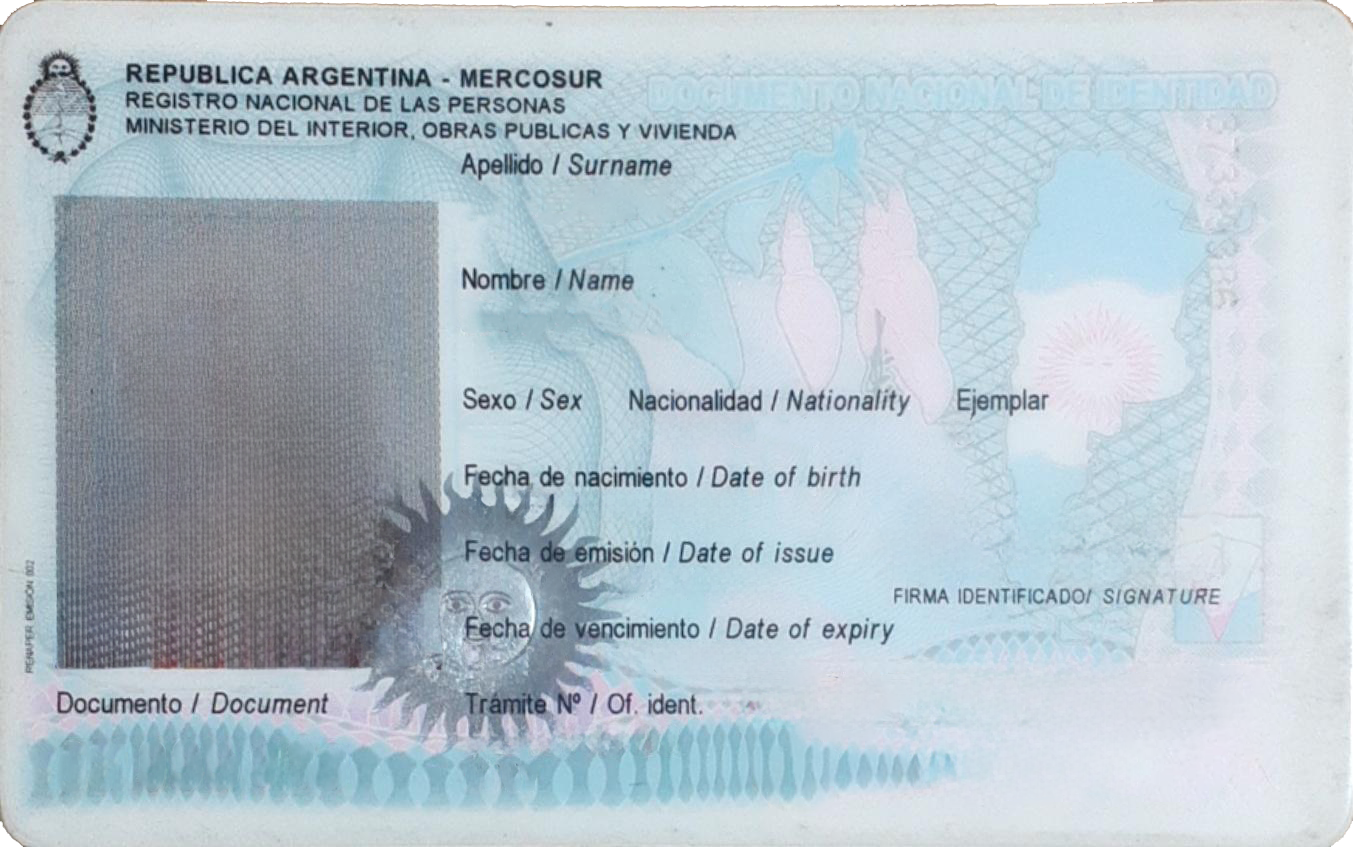}\label{fig:empty-arg}
}
\subfloat[CRI ID Card Template]{
  \includegraphics[width=35mm]{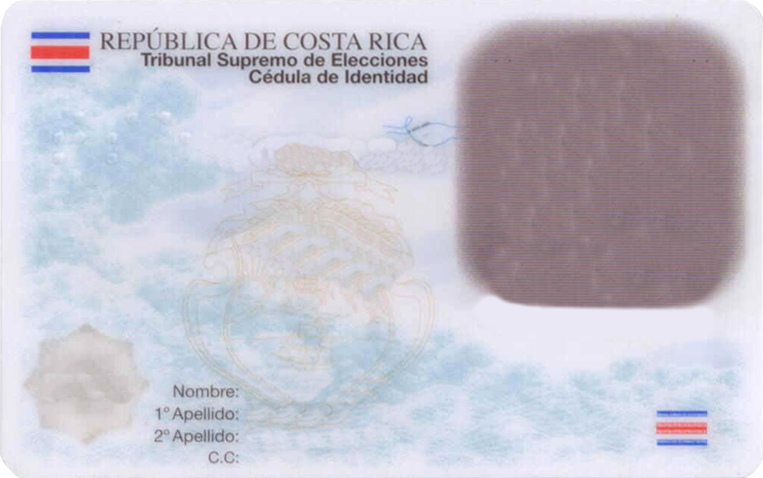}\label{fig:empty-cri}
}
\caption{Templates used as a background base for the creation of each sample for our different ID Card countries.}
\label{fig:examples-template-id}
\end{figure}

To create our identity cards, we began with voided front-facing digital templates of ID Cards from different countries, clearing all fields, signatures, and photographs using Adobe Photoshop\footnote{\url{https://www.adobe.com/es/products/photoshop.html}} to ensure a clean scenario without visual artefact, as shown in Figure~\ref{fig:examples-template-id}. This process allowed us to establish a base for generating a wide array of digital but realistic-looking ID Cards.

\begin{figure}[]
\centering
\subfloat[ESP ID Card Bona Fide]{
  \includegraphics[width=35mm,height=24mm]{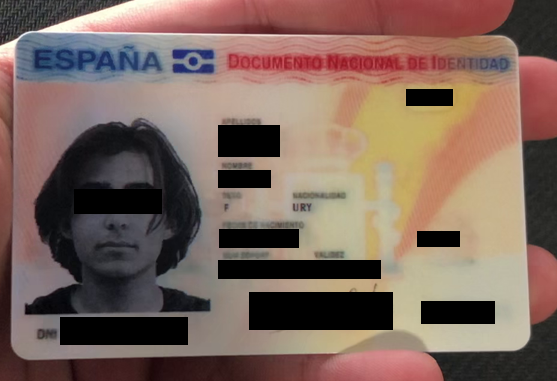}\label{fig:gen-esp}
}
\subfloat[ESP ID Card Screen Display]{
  \includegraphics[width=35mm,height=24mm]{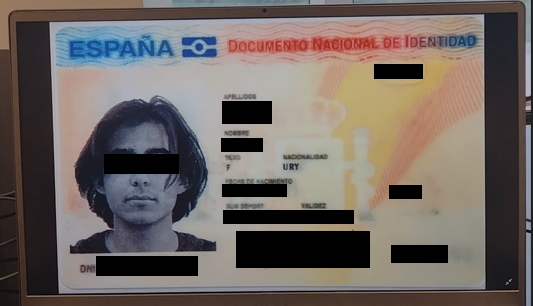}\label{fig:spoof-esp}
}

\subfloat[CHL ID Card Bona Fide]{
  \includegraphics[width=35mm,height=24mm]{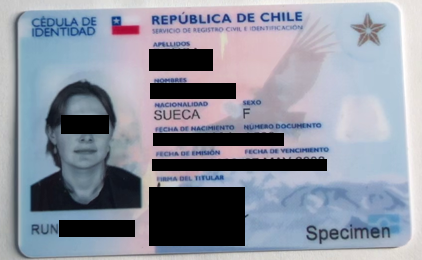}\label{fig:gen-chl}
}
\subfloat[CHL ID Card Screen Display]{
  \includegraphics[width=35mm,height=24mm]{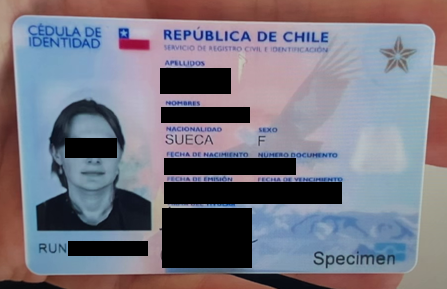}\label{fig:spoof-chl}
}

\subfloat[ARG ID Card Bona Fide]{
  \includegraphics[width=35mm,height=23mm]{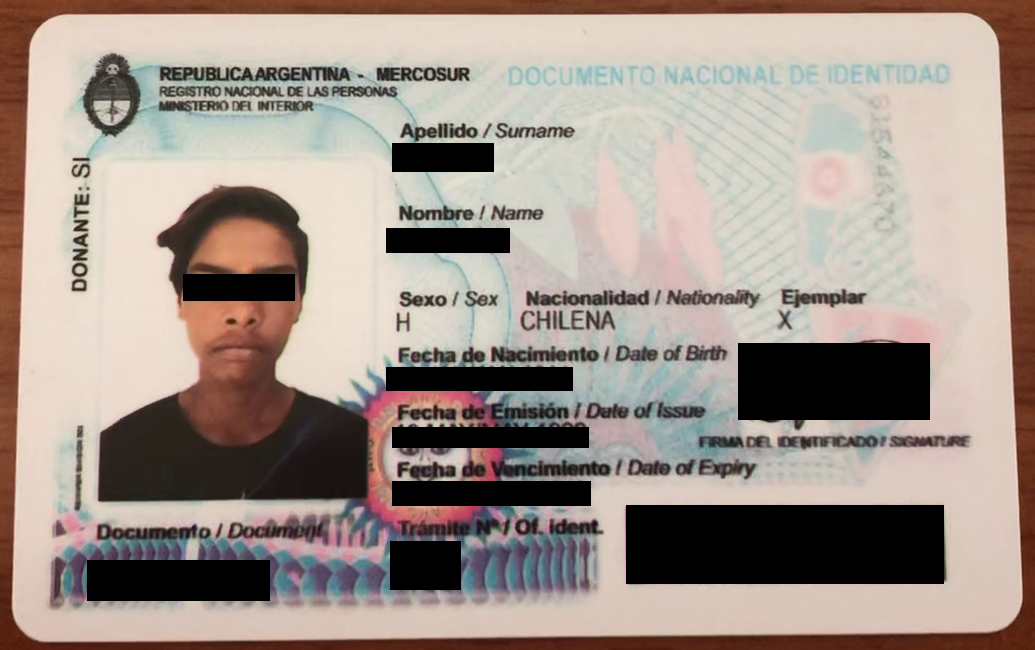}\label{fig:gen-arg}
}
\subfloat[ARG ID Card Screen Display]{
  \includegraphics[width=35mm,height=23mm]{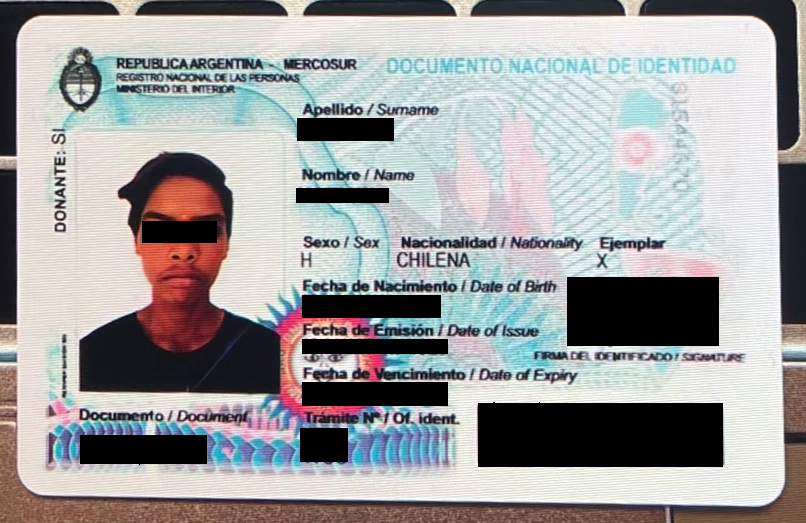}\label{fig:spoof-arg}
}

\subfloat[CRI ID Card Bona Fide]{
  \includegraphics[width=35mm,height=23mm]{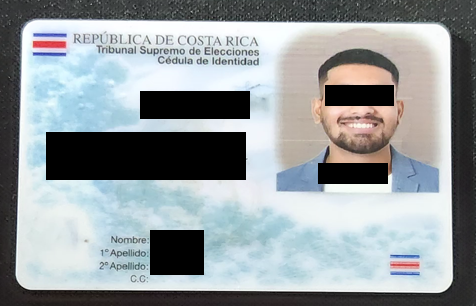}\label{fig:gen-cri}
}
\subfloat[CRI ID Card Screen Display]{
  \includegraphics[width=35mm,height=23mm]{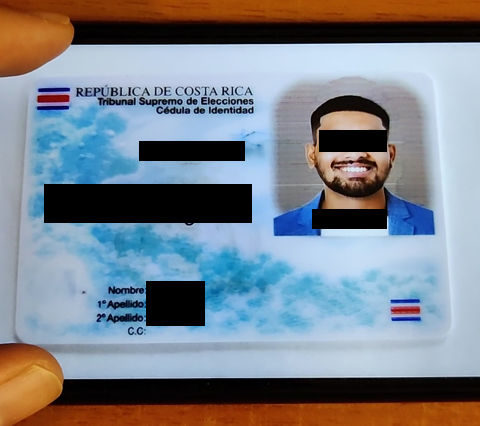}\label{fig:spoof-cri}
}
\caption{Examples of Fake ID Cards generated for the experiments.}
\label{fig:inputs-networks}
\end{figure}

Following the preparation of these templates, we employed a series of algorithms based on computer vision, such as Seamless Cloning to the entire images and also by blocks \cite{poisson}, designed to remove traces when new information is included. Further on that, random faces and signatures sourced from public datasets \cite{face-dataset-1, face-dataset-2, face-dataset-3, signature-dataset-1, signature-dataset-2, signature-dataset-3}, alongside the generation of plausible names, dates, and alphanumeric characters were used. This approach ensured that each digital ID Card was unique, embodying a realistic mix of data that could naturally occur on genuine identity documents. The result, in Figure~\ref{fig:inputs-networks}, was a diverse collection of our digital ID Cards, ready to be used in our proposal.

This approach allows us to overcome the limitations of public datasets and tailor our dataset to the specific requirements of our study. The private dataset, with its emphasis on digital ID Card representations from a diverse array of Latin American countries, positions us to address the critical gaps in current research and contribute valuable insights into detecting and preventing presentation attacks on ID Cards.

In summary, in our study, the bona fide class consists of homemade ID Cards that have been printed on PVC cards from digital templates. These physical cards represent bona fide IDs that one would typically encounter in real-world scenarios. Conversely, the attack display screen class is constituted by images of these same PVC-printed ID Cards, which are then displayed on various digital screens and subsequently recaptured. This process is designed to emulate screen display attacks, where a digital device is used to present a manipulated ID image, attempting to deceive ID verification systems. By displaying and capturing the bona fide PVC card IDs on different screens, we generate a dataset of spoofing attacks, each classified as an instance of the attack display screen class.

Thus, Figure~\ref{fig:inputs-networks} shows bona fide samples of the printed documents in PVC, and the same ID Card captured its displays on a screen source.

In Table~\ref{tab:data-dist}, the distributions by country, the number of unique users available, the number of screen sources, and the total number of images per country are detailed. The term ``screen sources" refers to the variety of different devices used to capture these document attacks, underscoring the breadth of our approach in simulating realistic scenarios where documents might be fraudulently presented through multiple digital screens. This diversity in screen sources is crucial for ensuring that our FSL models are trained on data that closely mimics the variations and nuances of real-world attacks.

\begin{table}[H]
\centering
\caption{Distribution of our dataset by country.}
\label{tab:data-dist}
\resizebox{\columnwidth}{!}{%
\begin{tabular}{@{}lccc@{}}
\toprule
\textbf{Country} & \textbf{Nº Users} & \textbf{Nº Screen Sources} & \textbf{Nº Images} \\ \midrule
Spain (ESP)            & 54                & 11                 & 23,504             \\
Chile (CHL)           & 30                & 9                  & 13,123             \\
Argentina (ARG)       & 30                & 9                  & 11,926             \\
Costa Rica (CRI)       & 30                & 5                  & 7,458              \\
 \bottomrule
\end{tabular}%
}
\end{table}
\vspace{-0.3cm}

\section{Metrics} 
\label{sec:metrics}  

The detection performance of the PAD biometric algorithms is standardised according to ISO/IEC 30107-3\footnote{\url{https://www.iso.org/standard/79520.html}}. The most relevant metrics for this study are Attack Presentation Classification Error Rate ($APCER$), Bona fide Presentation Classification Error Rate ($BPCER$), and $BPCER_{AP}$. Those metrics determine the error rates when classifying an instance between bona fide and the different Presentation Attack Instrument Species (PAIS).  

The $APCER$ metric measures the percentage of attack presentations incorrectly classified as bona fide for each PAIS. The worst-case scenario is considered when evaluating an entire system. The computation method is detailed in Equation~\ref{eq:apcer}, where the value of $N_{PAIS}$ corresponds to the number of attack presentation images, $RES_{i}$ is $1$ if the $i$th image is classified as an attack, or $0$ if it was classified as a bona fide presentation.
\vspace{-0.3cm}

\begin{equation}\label{eq:apcer}
    {APCER_{PAIS}}=1 - \frac{1}{N_{PAIS}}\sum_{i=1}^{N_{PAIS}}RES_{i}
\end{equation}

On the other hand, the $BPCER$ metric measures the proportion of bona fide presentations wrongly classified as attacks. The $BPCER$ can be computed using Equation~\ref{eq:bpcer}, where $N_{BF}$ is the amount of bona fide presentation images, and $RES_{i}$ takes the same values described in the $APCER$ metric. Together, the two metrics determine the system's performance, and they are subject to a specific operation point. 

\begin{equation}\label{eq:bpcer}
    BPCER=\frac{\sum_{i=1}^{N_{BF}}RES_{i}}{N_{BF}}
\end{equation} 

Finally, $BPCER_{AP}$ and the Equal Error Rate $(EER)$ are used to analyse the system performance on a specific operating point. The latter is the operating point where $APCER$ and $BPCER$ are equal. This operating point corresponds to the intersection with the diagonal line in a Detection Error Trade-off (DET) curve, which is also reported for all the experiments. On the other hand, the $BPCER_{AP}$ is the $BPCER$ value when the $APCER$ is $100/AP$. In this work, $BPCER_{10}$, $BPCER_{20}$ and $BPCER_{100}$ were evaluated, which correspond to $APCER$ values of 10\%, 5\% and 1\% respectively.

\section{Method}
\label{sec:method}

In our paper, we propose a method that takes advantage of the principles of Prototypical Networks for Few-shot Learning to address classification tasks with limited data. Our approach is centred around a selected set of support sets, which comprises a balanced array of samples representing both bona fide and display attack classes for each country included in our study. 

In a typical $N$-way $K$-shot scenario, this support set consists of two classes, each represented by $K$ examples that encapsulate the defining characteristics of that class.

The methodology begins with feature extraction, where each sample from the ``Support set'' (ID Cards from different countries) is processed through a neural network, acting as a feature extractor that converts raw data into an embedding vector. These representations are then used to calculate the centroid or mean embedding vector of the data points associated with each class, thereby establishing a prototype that epitomises the ``average'' example within that class's feature space.

Parallel to the support set, we have the ``Query set (Q)'', which includes examples that the model needs to classify. These query examples undergo the same feature extraction procedure to ensure they are comparable to the ``Support set (S)" in the embedding space. The classification process involves measuring the distance between the ``Query examples" and each class's prototype (C), typically using a Euclidean distance, and assigning the ``Query example" to the class of its nearest prototype. 

\begin{figure*}[]
    \centering
    \includegraphics[scale=0.23]{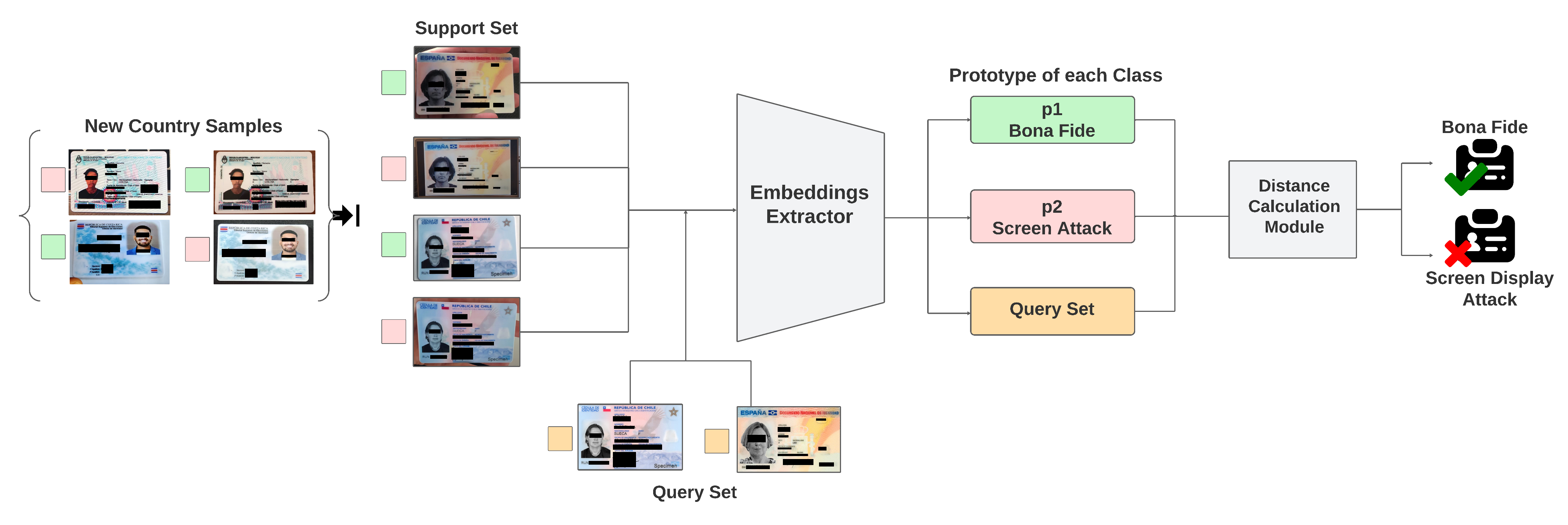}
    \caption{Proposed flow-based FSL method applied to the PAD Detection on the ID Card. The new country samples represent an unknown dataset.}
    \label{fig:prototypical-net-flow}
\end{figure*}

In Figure~\ref{fig:prototypical-net-flow}, we present the schematic workflow of the proposed method from our paper, illustrating the incorporation of Prototypical Networks for identity document verification. The diagram begins with a support set comprising samples from Chile and Spain, which the model uses as a foundation for learning. As our method evolves to enhance its adaptability and broaden its knowledge, we introduce FSL samples from new countries, specifically Costa Rica and Argentina, into the support set. This strategic addition allows the model to learn from a richer, more diverse dataset, imbuing it with the nuanced understanding necessary to authenticate ID Cards from a wider array of countries. Through this schematic representation, the paper effectively showcases the method's scalability and its capacity to integrate new data seamlessly into the FSL framework.

The learning process is an optimisation, with the model trained to minimise the distance between the query examples and their correct prototypes. Crucially, this process simulated a fine-tuning, where a model rapidly adjusts to new data by iterating through numerous training episodes, each with its own support and query set.

On the other hand, the loss function plays a crucial role in teaching the model to classify new data points correctly. The model starts by transforming the data from both the support and query sets into a high-dimensional embedding space. For every class, a central point or prototype is computed by finding the average of all the embedded points from the support set. 

When a new, unseen data point comes in from the query set, the model calculates how far this point is from each class's prototype. These distances are then used to estimate the likelihood of the query point belonging to each class, with the soft-max function converting the distances into probabilities. The closer a point is to a prototype, the higher the probability it is part of that class. The model's loss, calculated through Categorical Cross-Entropy, serves as a measure of the model's performance in correctly identifying the true class of each query point. 

The Categorical Cross-Entropy loss function is a crucial component in our FSL approach, particularly adapted for the classification of ID Cards. It quantifies the discrepancy between the predicted probabilities and the actual class labels, effectively guiding the model during the training process. By striving to minimise this loss, the model is effectively learning to pull query points closer to their true class prototypes and push them away from the prototypes of other classes, thereby continuously sharpening its classification capabilities. In order to clarify the process, a pseudocode is presented as follows in Algorithm ~\ref{alg:loss}.

\begin{algorithm}
\scriptsize
\caption{Loss calculation in our approach.} \label{alg:loss}
\begin{algorithmic}[1]
\STATE // Assume classes $C$ for ID Cards
\STATE // Assume support set $S$
\STATE // Assume query set $Q$
\STATE // Assume $f$ represents the embedding function
\FOR{each class $c$ in $C$}
    \STATE $S_c \leftarrow$ get support ID Card examples for class $c$ from $S$
    \STATE $\text{prototype}_c \leftarrow \frac{1}{|S_c|} \sum_{(x_i, y_i) \in S_c} f(x_i)$
\ENDFOR
\STATE $L \leftarrow 0$ // Initialise loss
\FOR{each ID Card example $(x, y)$ in $Q$}
    \STATE $p \leftarrow$ soft-max($-\text{distance}(f(x), \text{prototype}_y)$ for each prototype in $C$)
    \STATE $L \leftarrow L - \log(\text{prototype}_y)$ // Negative log probability for the true class
\ENDFOR
\STATE $L \leftarrow \frac{L}{|Q|}$ // Average loss over the query set
\STATE // Update model parameters to minimise $L$
\end{algorithmic}
\end{algorithm}

The iterative learning process unfolds over multiple epochs, each presenting the model with fresh class examples to ensure a robust learning trajectory. The sum of the loss through different points directs back-propagation, steering the model's parameters to usher the correct class prototypes closer to the query examples in the embedding landscape. Simultaneously, it distances the prototypes of incorrect classes, thus sharpening the model's classification performance. 

The model's feature extractor and embedding space are refined through repeated optimisation, allowing the network to understand class clustering better. This fosters an environment where even a few data points can guide precise classification decisions.

\section{Experiment and Results}
\label{sec:exp-result}

\begin{table*}[]
\centering
\scriptsize
\caption{Description of the composition of the experiments.}
\label{tab:exp-descrip}
\resizebox{\textwidth}{!}{%
\begin{tabular}{@{}ll|ccc|ccc|ccc@{}}
\toprule
                               &  & \multicolumn{3}{c|}{\textbf{Train Split}} & \multicolumn{3}{c|}{\textbf{Validation Split}} & \multicolumn{3}{c}{\textbf{Test Split}} \\ \midrule
\multicolumn{1}{l|}{\textbf{Exp. Name}} &
  \multicolumn{1}{c|}{\textbf{Countries}} &
  \textbf{Nº Users} &
  \textbf{Nº Screen} &
  \textbf{Nº Imgs} &
  \textbf{Nº Users} &
  \textbf{Nº Screen} &
  \textbf{Nº Imgs} &
  \textbf{Nº Users} &
  \textbf{Nº Screen} &
  \textbf{Nº Imgs} \\ \midrule
\multicolumn{1}{l|}{Exp. 1 - Baseline} & ESP - CHL & 48          & 11         & 19,696         & 16           & 11           & 6,241           & 21         & 11         & 9,741        \\
\multicolumn{1}{l|}{Exp. 2}    & ESP - CHL - CRI & 53 & 11         & 20,022         & 21           & 11            & 20,322          & 26         & 11          & 13,193        \\
\multicolumn{1}{l|}{Exp. 3}    &ESP - CHL - ARG  & 53          & 11          & 19,751          & 21           & 11            & 6,850           & 26         & 11         & 16,563        \\ \bottomrule
\end{tabular}
}
\end{table*}

\subsection{Experiment 1: Baseline}
Three experiments were developed in order to show the performance of our proposal. Throughout the development of our experiment, numerous networks were evaluated to devise the most effective approach for distinguishing between bona fide ID Cards and screen display presentation attacks. The best configuration is based on EfficientNetV2-B0 architecture \cite{efficientnetv2}. This choice was motivated by its balance between efficiency and low EER, leveraging the initial weights from ImageNet1K \cite{deng2009imagenet} to ensure a robust starting learning point.

This research uses all available ID Cards from Spain and Chile and splits them into train, validation, and test sets to support a comprehensive evaluation framework. The data split, broken down into images, unique users, and screen sources, can be seen in Table~\ref{tab:exp-descrip}. 

Image inputs were aligned, zero-padded to maintain an aspect ratio without distortion, and restricted to the front side of the ID for consistency and focus. After extensive testing and comparison against alternatives, this configuration demonstrated the best performance, indicating its suitability as a robust baseline for identifying presentation attacks in ID document verification tasks. 
The data augmentation~\cite{imgaug} made to the document images was a small variation to the bounding box of the document in the image to give it some variability before cropping the document. 

It is important to note that while data augmentation techniques are often effective in enhancing model performance by introducing variability and robustness to the training process, in this specific task, such techniques did not yield the expected improvements. Despite experimenting with several functions of data augmentation, including changes in lighting, blur, colour temperature adjustments, horizontal flipping, and JPEG compression, the results did not show any significant enhancement in the model's ability to distinguish between presentation attacks and bona fide ID Cards. This outcome suggests that the specific challenge of identifying screen display attacks and bona fide ID Cards may not be effectively captured or emphasised through traditional data augmentation methods.

The FSL preprocessing strategy to calculate the prototypes was set to ``Average'', which means the embedding average of each prototype (ID Card per country) optimised the way we prepared our data for input into the model. 
Euclidean distance was used to calculate the distance between the new samples and the prototypes of the known classes. 

Our query set consisted of 42 subject images, balanced to include an equal number of screen display attacks and bona fide images, ensuring fairness and challenge in the model's evaluation capabilities. 

The support set consisted of 8 subject images, equally divided between IDs from Spain (ESP) and Chile (CHL). Each country's contributions were evenly split between screen attack and bona fide samples. This model was trained to binary classify inputs as either $0$ (screen display presentation attack) or $1$ (bona fide), providing a clear and direct output for analysis.

To fine-tune the model's learning process, a learning rate ($lr$) of $5e^{-4}$ was selected in conjunction with the AdamW optimiser, which included a weight decay ($wd$) of $1e^{-5}$. This combination was identified as optimal for our purposes, balancing the rate of learning with the need to prevent overfitting based on a grid-search. Moreover, an early stopping mechanism was implemented with the patience of $30$ epochs, safeguarding against unnecessary computational expense by halting training when improvement plateaued.

Table \ref{tab:resume-results} shows a summary of the results for all the experiments. All the results are in percentages. For experiment 1, The EER and BPCER10, BPCER20, and BPCER100 depicted the lower generalisation capabilities to new countries such as Costa Rica and Argentina in the traditional approach without FSL. 

\begin{figure*}[]
    \centering
    \includegraphics[scale=0.44]{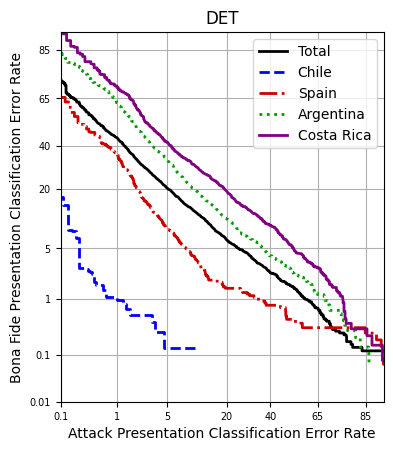}
    \includegraphics[scale=0.44]{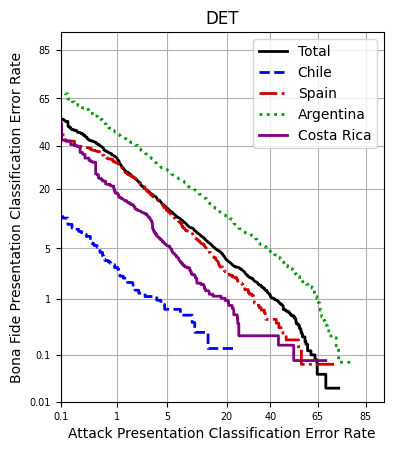}
    \includegraphics[scale=0.44]{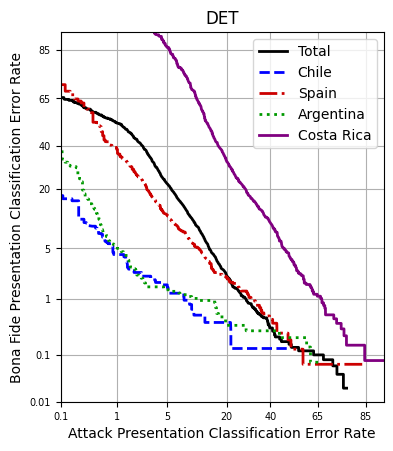}
    \caption{DET curves by country. Left. Experiment 1, Middle: Experiment 2, and Right: Experiment 3.}
    \label{fig:det-exp-1}
\end{figure*}

Figure~\ref{fig:det-exp-1} left, presents a DET curve that illustrates the performance of Experiment 1, delineating its efficacy in distinguishing between bona fide and attack presentations for Chile and Spain. The plot also provides a visual comparison of the algorithm's performance metrics, specifically for Spain and Chile and the worse results for Argentina and Costa Rica. Each country's curve offers insight into the balance between APCER and BPCER achieved by our model, reflecting the particularities of the algorithm's performance in each geographic context. 

\begin{table*}[]
\centering
\scriptsize
\caption{Summary results for all the experiments}
\label{tab:resume-results}
\resizebox{\textwidth}{!}{%
\begin{tabular}{@{}lcccc|cccc|cccc@{}}
\toprule
 &
  \multicolumn{4}{c|}{\textbf{Experiment 1: Baseline - Testing Set (\%)}} &
  \multicolumn{4}{c|}{\textbf{Experiment 2 - Testing Set (\%)}} &
  \multicolumn{4}{c}{\textbf{Experiment 3 - Testing Set (\%)}} \\ \midrule
 &
  \textbf{Spain} &
  \textbf{Chile} &
  \textbf{Costa Rica} &
  \textbf{Argentina} &
  \textbf{Spain} &
  \textbf{Chile} &
  \textbf{Costa Rica} &
  \textbf{Argentina} &
  \textbf{Spain} &
  \textbf{Chile} &
  \textbf{Costa Rica} &
  \textbf{Argentina} \\ \midrule
EER      & 6.54  & 0.97 & 19.52 & 14.64 & 8.16 & 1.74 & 5.32 & 14.55 & 7.76  & 2.30 & 24.12 & 2.11 \\

BPCER10  & 4.01  & 0.14 & 30.25 & 19.68 & 6.77 & 0.41 & 2.27 & 19.89  & 5.59 & 0.54 & 68.10 & 1.01 \\

BPCER20  & 8.22  & 0.14 & 41.61 & 32.78 & 12.82 & 0.68 & 5.41 & 27.85  & 11.37 & 1.49 & 85.81 & 1.37 \\

BPCER100 & 34.91 & 0.95 & 70.21 & 62.08 & 31.03 & 2.85 & 18.42 & 46.23 & 35.96 & 4.20 & 95.68 & 4.99 \\ \bottomrule
\end{tabular}%
}
\end{table*}

\subsection{Experiment 2: FSL with Costa Rica images}

According to our previous experiments with the EfficientNetV2-B0 architecture and Prototypical Networks for FSL, this new iteration maintains the previous configuration while introducing further diversity in the dataset to challenge and enhance the model's capabilities. The base dataset remains consistent with our prior work, featuring ID Cards from Spain and Chile, ensuring a steady foundation for comparative analysis across experiments.

In this iteration, we extend our exploration into the adaptability and efficacy of our model by incorporating a selection of ID Card users from Costa Rica, thereby adding a new dimension of geographic diversity. Specifically, the training set is enriched with 5 unique users from Costa Rica, providing fresh data for the model to learn from. The validation set follows suit, including another distinct group of 5 unique Costa Rican users, allowing us to rigorously assess the model's ability to generalise to entirely unseen data within the same framework. The data split, broken down into images, unique users, and screen sources, can be seen in Table~\ref{tab:exp-descrip}.

A notable enhancement in this experiment is the increase in the number of images per screen source from Costa Rica, with each contributing 15 images to the train set. This significant increase in data volume from a single country aims to deepen our understanding of how well our model can perform under conditions of both limited user diversity and increased sample size per user.

Regarding the support set was constructed to offer a balanced assortment of ID Card images from Spain, Chile, and Costa Rica, incorporating an equitable mix of bona fide and fake examples from each country. The results of our algorithm's performance, delineated by individual country, are tabulated in Table~\ref{tab:resume-results}. Enhancing the insights provided by these results.

Figure~\ref{fig:det-exp-1} middle presents a DET curve that illustrates the performance of Experiment 2, which visually captures the precision of our algorithm in distinguishing between bona fide and Fake ID Card representations when a new country (Costa Rica) was added to the FSL.

\subsection{Experiment 3: FSL with Argentine images}

Building on the framework of our previous experiments, which showcased the efficiency of the EfficientNetV2-B0 architecture with Prototypical Networks in FSL, we explored another iteration with a similar setup but introduced a new country in the dataset composition.

As with earlier trials, the foundational dataset comprised ID Cards from Spain and Chile, maintaining consistency in the data's geographical origin. However, this experiment introduced a distinct variation by incorporating users from Argentina, adding another layer of diversity to test the model's adaptability and performance.

In this specific setup, the training set included five unique users from Argentina to introduce new patterns and characteristics for the model to learn from. Similarly, the validation set was enriched with another set of Five unique Argentine users, distinct from those in the training set, to evaluate the model's generalisation capabilities across unseen data. Each screen source in this experiment contributed 15 images. The data split, broken down into images, unique users, and screen sources, can be seen in Table~\ref{tab:exp-descrip}.
Finally, the support set has been thoughtfully assembled to maintain an even distribution of ID Card images from Spain, Chile, and Argentina, covering both authentic and counterfeit instances for each category. The algorithm's performance metrics, segregated by these specific countries, are comprehensively documented in Table~\ref{tab:resume-results}. To further show the effectiveness of our approach.

Figure~\ref{fig:det-exp-1} right, presents a DET  curve that illustrates the performance of Experiment 3, which visually captures the precision of our algorithm in distinguishing between bona fide and Fake ID Card representations when a second new country (Argentina) was added to the FSL. 
Table \ref{tab:resume-results} right, shows the results of Experiment 3 in percentages. The EER and BPCER10, 20, and 100 depicted an improvement in generalisation capabilities for new countries, especially in Argentina now. In the case of Costa Rica ID Cards, even if we did not include any samples in the training process, the performance is still competitive.

\section{Conclusion}
\label{sec:conclu}

Our paper has explored the efficacy of FSL in the context of ID Card verification, its ability to achieve high performance with a relatively small number of images, and its generalisation to new countries. Our research has demonstrated that in Prototypical Networks based on EfficientNetV2-BO, we can attain very competitive performance metrics even with limited data. 

\section*{Acknowledgements}
This research has been partially funded by Facephi, the European Union’s Horizon 2020 research and innovation program under grant agreements No 883356, 101121280 and the German Federal Ministry of Education and Research, and the Hessian Ministry of Higher Education, Research, Science, and the Arts, which jointly support the National Research Center for Applied Cybersecurity ATHENE.

{\small
\bibliographystyle{ieee}
\bibliography{bibilography}
}

\end{document}